\documentclass[11pt]{article}

\usepackage[margin=1in]{geometry}
\usepackage[T1]{fontenc}
\usepackage[utf8]{inputenc}
\usepackage{lmodern}
\usepackage{microtype}
\usepackage{amsmath,amssymb,amsthm}
\usepackage{enumitem}
\usepackage{graphicx}
\graphicspath{{figures/}}
\usepackage{booktabs}
\usepackage{authblk}
\usepackage{caption}
\usepackage{subcaption}
\usepackage[authoryear,round]{natbib}
\usepackage[hidelinks]{hyperref}

\theoremstyle{plain}
\newtheorem{theorem}{Theorem}[section]

\newtheorem{corollary}[theorem]{Corollary}
\newcommand{\energyunit}{\ensuremath{\mathrm{kcal\,mol^{-1}}}}
\newcommand{\forceunit}{\ensuremath{\mathrm{kcal\,mol^{-1}\,\mathring{A}^{-1}}}}
\title{AdaptNTK: Adaptive Uncertainty Quantification and Active Learning for Neural Network Potentials}
\author[1]{Prajwal Ananth}
\author[2,*]{Shuwen Yue}

\affil[1]{Center for Applied Mathematics, Cornell University, Ithaca, NY, USA}
\affil[2]{R. F. Smith School of Chemical and Biomolecular Engineering, Cornell University, Ithaca, NY, USA}
\affil[*]{Corresponding author: 
\href{mailto:shuwen.yue@cornell.edu}{shuwen.yue@cornell.edu}}

\date{}

\begin{document}

\maketitle

\begin{abstract}
Machine learning interatomic potentials bridge the gap between quantum chemical precision and classical computational speed, enabling molecular dynamics simulations with first-principles accuracy. Their reliability is often improved through active learning, which iteratively expands the training set by identifying uncertain, out-of-distribution configurations. Existing uncertainty-quantification methods often involve a trade-off between computational cost and reliability, and generally cannot account for redundancy as an acquisition batch is assembled. Here, we introduce AdaptNTK, a single-model framework that measures uncertainty as a regularized Mahalanobis distance in empirical neural tangent kernel (NTK) feature space. With the NTK features fixed during acquisition, the uncertainty depends on the acquired configurations but not their reference labels. This allows the uncertainty to be updated recursively after each selection without retraining, reducing redundancy within an acquisition batch. On held-out rMD17 data, AdaptNTK achieves the highest mean correlations with force errors (Spearman $0.68$, Pearson $0.71$) and matches a three-member ensemble in error retention. In active learning experiments, AdaptNTK achieves the lowest force errors across rMD17 and Transition-1X, with particularly strong performance on transition-state configurations in Transition-1X. AdaptNTK is $2.6\times$ faster per Transition-1X cycle than the ensemble, providing efficient single-model uncertainty estimation with sequential updates for data-efficient active learning.
\end{abstract}

\section{Introduction}
\label{sec:introduction}
Molecular dynamics (MD) simulations are widely used to study chemical processes at atomistic resolution. Their accuracy and range, however, have traditionally been limited by a trade-off between computational cost and the quality of the underlying potential-energy surface (PES). Electronic-structure methods accurately describe bond breaking, charge redistribution, and small energy differences but remain computationally costly, whereas empirical force fields can access larger length and timescales but often do not generalize to new chemical environments. Machine learning interatomic potentials (MLIPs) address this limitation by learning potentials from reference calculations and approximating first-principles energies and forces at computational costs that can scale linearly with system size~\citep{behler2007, bartok2010gap, unke2021review, batzner2022nequip, batatia2022mace, musaelian2023allegro}. 

Despite their efficiency, MLIPs are strongly dependent on their training data. They generally interpolate well within regions of configuration space represented in the training set, but their predictions can deteriorate when trajectories encounter configurations outside these regions. Such extrapolation errors can lead to inaccurate forces and, consequently, incorrect trajectories. Reliable uncertainty estimates are therefore important for identifying configurations for which model predictions may not be trustworthy. 

Gaussian-process (GP) potentials naturally provide uncertainty estimates through their posterior variance, which depends on the training configurations but not their reference labels. This enables uncertainty to be evaluated and updated analytically as the training set changes. Neural-network potentials, on the other hand, are more expressive but generally lack comparable closed-form uncertainty estimates. Their uncertainty is typically approximated using ensembles~\citep{seung1992qbc, smith2018less} or approximate Bayesian methods~\citep{gal2016dropout, maddox2019swag, amini2020evidential, soleimany2021evidential}, which can be computationally expensive or unreliable outside the training distribution. Developing reliable and efficient uncertainty estimates for a single trained MLIP therefore remains an important challenge. 

Active learning is a natural downstream application of uncertainty quantification. Configurations with high uncertainty can be selected for reference labeling and added to the training set~\citep{podryabinkin2017, smith2018less, vandermause2020flare, zhang2020dpgen}. When multiple configurations are selected in a batch, however, uncertainty estimates computed only once at the beginning of the batch do not account for configurations that have already been selected, potentially leading to redundant acquisitions. GP posterior variance naturally supports updating the uncertainty after each selection without requiring the corresponding reference label~\citep{vandermause2020flare}. Achieving a similar capability for neural-network potentials is more challenging, because conventional uncertainty estimates generally require additional model fitting or retraining to incorporate newly selected configurations.  Parameter-gradient features associated with the neural tangent kernel (NTK)~\citep{jacot2018} provide a principled framework for recovering this GP-like capability in trained neural-network potentials.

Here we introduce AdaptNTK, a framework for uncertainty quantification and active learning in neural-network potentials. AdaptNTK represents each configuration using the gradient of its predicted energy with respect to all model parameters and assigns a regularized distance from the training data in this tangent-feature space. Because the uncertainty depends on the selected configurations but not their reference labels, it can be updated analytically after each selection without retraining the network while the acquisition batch is being assembled. Our contributions are as follows:

\begin{itemize}[leftmargin=*,nosep]
    \item We connect regularized empirical-NTK uncertainty to feature-space geometry, evaluate its ranking, retention, and calibration, and prove a Gaussian-sketch guarantee controlled by the ridge effective dimension.
    \item We use a label-free rank-one update to condition the remaining uncertainty scores after each selection, reducing within-batch redundancy without additional model retraining.
    \item  We benchmark AdaptNTK's force-error ranking, error retention, uncertainty calibration, and active-learning performance against ensemble and single-model baselines on rMD17~\citep{christensen2020rmd17} and Transition-1X~\citep{schreiner2022transition1x}.
\end{itemize}

Overall, AdaptNTK achieves the highest mean correlations with force error among the evaluated uncertainty quantification methods and the lowest force errors in the active-learning experiments on rMD17 and Transition-1X. Its normalized risk--coverage performance is comparable to that of a three-member ensemble, while its wall-clock cost is less than half that of the ensemble in the reported Transition-1X timing experiment. Together, these results show that sketched empirical-NTK uncertainty provides a practical single-model foundation for data-efficient active learning of neural-network potentials.

\section{Related Work}
\label{sec:related-work}
Several uncertainty-quantification methods have been developed
for estimating predictive uncertainty in deep neural networks~\citep{seung1992qbc,
gal2016dropout, lakshminarayanan2017ensembles, maddox2019swag, amini2020evidential,
soleimany2021evidential}. These approaches have also been adapted to atomistic
machine learning, particularly through ensembles and
query-by-committee (QBC) methods~\citep{smith2018less, podryabinkin2017,
zhang2020dpgen, kahle2022quality}. More recent work has developed single-model
or reduced-cost alternatives based on dropout and learned-feature
distributions~\citep{wen2020dropout, zhu2023gmm, bigi2024llpr,
chong2025rigidity}, evidential regression~\citep{xu2026eip}, shared-backbone
ensembles~\citep{kellner2024shallow, beck2025multihead}, variational
inference~\citep{coscia2026blips, mamun2026dgkl}, and post-hoc uncertainty estimation and
calibration~\citep{vita2025ltau, perez2025misspecified, ho2026calibration}.
However, these approaches often involve a trade-off between computational
scalability and the reliability of their uncertainty
estimates~\citep{tan2023singlemodel, bilbrey2025foundation}. More recently, \citet{wilson2025uncertainty} showed that uncertainty quantification based on empirical neural tangent kernels (NTKs) can match or outperform deep ensembles at substantially lower computational cost in general regression settings. Their approach is post-hoc and sampling-based, constructing ensembles through gradient-descent sampling of linearized networks, whereas our uncertainty estimate is obtained directly from a closed-form regularized quadratic form in empirical NTK feature space.

Batch active-learning methods have also been extensively studied across machine-learning applications, including image classification and tabular regression. Last-layer and gradient-feature representations are commonly paired with diversity criteria for batch acquisition~\citep{holzmuller2023framework, ash2020badge, sener2018coreset}, with related approaches developed for interatomic potentials~\citep{zaverkin2022batch}. Concurrent work by \citet{vargaumbrich2026pretrained} and \citet{vargaumbrich2026forcentk} also uses neural tangent kernel features and diversity-aware selection for active learning in MLIPs. These works demonstrate the utility of feature-based representations and within-batch diversification for selecting informative training data. Their primary focus, however, is fixed-budget acquisition, where feature scores are used to rank candidates rather than to provide a quantitative, calibrated uncertainty for each configuration.  Our focus is complementary: we use empirical NTK features to construct a pointwise uncertainty measure that can be calibrated and updated adaptively as configurations are acquired. This distinction is important for molecular simulations, where uncertainty needs to be evaluated on individual configurations as the trajectory evolves. Such estimates can identify departures from the training distribution, provide stopping criteria for data acquisition, and indicate when model predictions may not be reliable during long-running simulations. 

\section{Background}
\label{sec:background}

\subsection{Gaussian processes}
\label{sec:background-gp}
A Gaussian Process (GP) defines a prior over functions $f : \mathbb{R}^d \to \mathbb{R}$ with mean function $m(\cdot)$ and covariance kernel $k(\cdot, \cdot)$~\citep{rasmussen2006gp}. Given a dataset $\mathcal{D} = (\mathbf{X}, \mathbf{y})$ with observation noise $\sigma^2$, the posterior predictive distribution at a test input $\mathbf{x}^\star$ is Gaussian with variance
\[
\Sigma(\mathbf{x}^\star) = k(\mathbf{x}^\star, \mathbf{x}^{\star}) - k_{\mathbf{x}^\star, \mathbf{X}} \left(\mathbf{K}_{\mathbf{X},\mathbf{X}} + \sigma^2 \mathbf{I}\right)^{-1} k_{\mathbf{X}, \mathbf{x}^{\star}}
\]
In this work, we focus on the predictive variance, which serves as a measure of uncertainty and forms the basis for our acquisition strategy.

\subsection{Neural tangent kernels}
\label{sec:background-ntk}
Consider a neural network $f(\cdot;\boldsymbol{\theta}) :
\mathbb{R}^d \to \mathbb{R}$ parameterized by
$\boldsymbol{\theta}\in\mathbb{R}^P$. Its neural tangent feature map at
parameters $\boldsymbol{\theta}$ is
$
\phi_{\boldsymbol{\theta}}(\mathbf{x})
=
\nabla_{\boldsymbol{\theta}} f(\mathbf{x};\boldsymbol{\theta})
\in \mathbb{R}^P,
$
which induces the empirical neural tangent kernel (NTK)
$
\mathcal{K}_{\boldsymbol{\theta}}(\mathbf{x},\mathbf{x}')
=
\phi_{\boldsymbol{\theta}}(\mathbf{x})^\top
\phi_{\boldsymbol{\theta}}(\mathbf{x}').
$
In the infinite-width limit, the NTK converges to a deterministic kernel, and,
under standard assumptions, neural-network training can be described by the
corresponding kernel dynamics~\citep{jacot2018,lee2019wide}.

Here, we instead use the empirical NTK of a finite, trained network. Treating
this kernel as a Gaussian-process covariance yields a predictive uncertainty that can be written directly in terms of the tangent-feature matrix
$\mathbf{\Phi}\in\mathbb{R}^{n\times P}$. This construction corresponds to a local linearization around the learned parameters and does not assume that the model operates in the infinite-width regime.

\section{Methods}
\label{sec:methods}

\subsection{Motivation and theoretical analysis}
\label{sec:methods-theory}

We first develop a geometric interpretation of predictive uncertainty by relating Gaussian-process (GP) variance induced by the empirical neural tangent kernel (NTK) to distances in NTK feature space.

Let $\mathbf{\Phi} \in \mathbb{R}^{n \times P}$ denote the empirical NTK feature matrix for the training dataset, where each row corresponds to the tangent feature representation of a training configuration. Treating the empirical NTK as the covariance kernel of the locally linearized neural network, the predictive variance at a query point $\mathbf{x}$ is given by
\begin{equation}
\label{eq:gp-variance-data-space}
    \mathbb{V}[f(\mathbf{x})]
    =
    \phi(\mathbf{x})^\top \phi(\mathbf{x})
    -
    \phi(\mathbf{x})^\top
    \mathbf{\Phi}^\top
    \left(
        \mathbf{\Phi}\mathbf{\Phi}^\top
        +
        \sigma^2 \mathbf{I}_n
    \right)^{-1}
    \mathbf{\Phi}
    \phi(\mathbf{x}),
\end{equation}
where $\sigma^2 > 0$ denotes the GP observation-noise or regularization parameter.

While Eq.~\eqref{eq:gp-variance-data-space} is expressed in the data space, it can be expressed in an equivalent formulation in the parameter space. Using the Woodbury identity in its push-through form, we rewrite the predictive variance as
\begin{equation}
\label{eq:gp-variance-parameter-space}
    \mathbb{V}[f(\mathbf{x})]
    =
    \phi(\mathbf{x})^\top
    \left[
        \mathbf{I}_P
        -
        \mathbf{F}
        \left(
            \mathbf{F}
            +
            \sigma^2 \mathbf{I}_P
        \right)^{-1}
    \right]
    \phi(\mathbf{x}),
\end{equation}
where $\mathbf{F}
    =
    \mathbf{\Phi}^\top \mathbf{\Phi}
    \in
    \mathbb{R}^{P \times P}$ is the empirical feature covariance matrix.

Equation~\eqref{eq:gp-variance-parameter-space} motivates a direct geometric interpretation in the interpolative limit $\sigma^2 \to 0^+$. In this limit,
\begin{equation}
\label{eq:training-feature-projector}
    \mathbf{F}
    \left(
        \mathbf{F}
        +
        \sigma^2 \mathbf{I}_P
    \right)^{-1}
    \longrightarrow
    \mathbf{P}_{\parallel}
    =
    \mathbf{F}^{+}\mathbf{F},
\end{equation}
where $\mathbf{F}^{+}$ denotes the Moore--Penrose pseudoinverse and $\mathbf{P}_{\parallel}$ is the orthogonal projector onto the subspace spanned by the training features. Consequently, the residual operator converges to the orthogonal projector onto the complementary subspace, $\mathbf{P}_{\perp}
    =
    \mathbf{I}_P
    -
    \mathbf{P}_{\parallel}.$ Substituting this limiting form into Eq.~\eqref{eq:gp-variance-parameter-space} gives
\[
    \lim_{\sigma^2 \to 0^+}
    \mathbb{V}[f(\mathbf{x})]
    =
    \phi(\mathbf{x})^\top
    \mathbf{P}_{\perp}
    \phi(\mathbf{x})
    =
    \left\|
        \mathbf{P}_{\perp}
        \phi(\mathbf{x})
    \right\|_2^2.
\]
Thus, in the noiseless limit, GP predictive uncertainty corresponds to the squared Euclidean distance of the query feature $\phi(\mathbf{x})$ from the subspace spanned by the training features in NTK feature space. A query whose tangent representation lies entirely within this subspace has vanishing uncertainty, whereas components orthogonal to the training-feature subspace contribute directly to the predictive variance.

For any finite $\sigma^2 > 0$, Eq.~\eqref{eq:gp-variance-parameter-space} can equivalently be written as
\begin{equation}
\label{eq:gp-variance-mahalanobis}
    \mathbb{V}[f(\mathbf{x})]
    =
    \sigma^2
    \phi(\mathbf{x})^\top
    \left(
        \mathbf{F}
        +
        \sigma^2 \mathbf{I}_P
    \right)^{-1}
    \phi(\mathbf{x}).
\end{equation}
The quadratic form in Eq.~\eqref{eq:gp-variance-mahalanobis} can be interpreted as a regularized squared Mahalanobis norm of the feature representation $\phi(\mathbf{x})$, with the metric determined by the regularized empirical feature covariance matrix. This provides a continuous extension of the subspace-distance interpretation above: feature-space directions that are weakly represented by the training data contribute more strongly to the uncertainty, whereas directions that are well represented are increasingly suppressed.

Motivated by this form, we define the empirical NTK uncertainty as
\begin{equation}
\label{eq:ntk-uncertainty}
    U(\mathbf{x})
    =
    \lambda
    \phi(\mathbf{x})^\top
    \left(
        \mathbf{F}
        +
        \lambda \mathbf{I}_P
    \right)^{-1}
    \phi(\mathbf{x}),
\end{equation}
where $\lambda > 0$ is a tunable regularization parameter. For conservative neural-network potentials, we take $\phi(\mathbf{x})
    =
    \nabla_{\theta}
    E(\mathbf{x};\theta)$
where $E(\mathbf{x};\theta)$ denotes the predicted energy and $\theta$ the model parameters. Since forces are obtained by differentiating the predicted energy with respect to atomic coordinates, these energy-gradient features encode the structure-dependent sensitivity of the learned potential-energy surface to perturbations in parameter space. Although they do not directly represent force errors, we empirically find that the resulting uncertainty correlates strongly with force prediction error.

Direct evaluation of Eq.~\eqref{eq:ntk-uncertainty} in the full parameter space is computationally expensive because $\phi(\mathbf{x}) \in \mathbb{R}^{P}$ is high-dimensional for over-parametrized neural-network potentials. We therefore introduce a randomized sketch that projects the NTK feature map into a lower-dimensional space while approximately preserving the geometry relevant to the uncertainty score.

\begin{theorem}[Regularization-aware preservation of NTK uncertainty]
\label{thm:ntk-sketch-preservation}
Let $\Phi\in\mathbb{R}^{n\times P}$ have rows $\phi(x_i)^\top$, let
$F=\Phi^\top\Phi$, and let $\mathcal X$ be a fixed set of $m$ queries. For each
$x\in\mathcal X$, let $Z_x$ be $\Phi$ with the row $\phi(x)^\top$ appended, and
define
\[
d_\lambda=\max_{x\in\mathcal X}\operatorname{tr}\!\left[Z_xZ_x^\top
\bigl(Z_xZ_x^\top+\lambda I_{n+1}\bigr)^{-1}\right].
\]
Let $S\in\mathbb{R}^{p\times P}$ have independent
$S_{ij}\sim\mathcal N(0,1/p)$, set $\widetilde\phi(x)=S\phi(x)$ and
$\widetilde F=SFS^\top$, and define
$\widetilde U(x)=\lambda\widetilde\phi(x)^\top
(\widetilde F+\lambda I_p)^{-1}\widetilde\phi(x)$. There is an absolute
constant $C$ such that, for $0<\epsilon<1$, if
\[
p\ge C\epsilon^{-2}\!\left(d_\lambda+\log\frac{m}{\delta}\right),
\]
then with probability at least $1-\delta$, simultaneously for all
$x\in\mathcal X$,
\[
|\widetilde U(x)-U(x)|\le\epsilon\bigl(\lambda+U(x)\bigr).
\]
\end{theorem}

Theorem~\ref{thm:ntk-sketch-preservation} shows that, for a fixed set of queries, a Gaussian sketch of dimension controlled by the ridge effective dimension is sufficient to preserve the full-space NTK uncertainty uniformly, up to a regularization-aware error of order $\epsilon(\lambda+U(x))$. The effective dimension is a sum of shrinkage factors $s_j^2/(s_j^2+\lambda)$ and may be far below both $P$ and the feature rank. This allows $U(x)$ to be evaluated in a substantially lower-dimensional space while closely preserving the full-space uncertainty.

\begin{corollary}[AdaptNTK decision stability]
\label{cor:adaptntk-decision-stability}
For a $B$-step full-space greedy path, augment $\Phi$ with previously selected
features and let $d_\lambda^{\rm AL}$ be the maximum ridge effective dimension
over all steps and remaining candidates. Replace $m$ by $Bm$ and $d_\lambda$ by
$d_\lambda^{\rm AL}$ in Theorem~\ref{thm:ntk-sketch-preservation}. If $U_t^{(1)}$ and
$U_t^{(2)}$ are the two largest full-space scores at step $t$ and
\[
\min_{t<B}\frac{U_t^{(1)}-U_t^{(2)}}
{2\lambda+U_t^{(1)}+U_t^{(2)}}>\epsilon,
\]
then the sketched and full-space rules select the same ordered batch.
\end{corollary}

Corollary~\ref{cor:adaptntk-decision-stability} gives the corresponding active-learning consequence. Once the sketch error is smaller than the relevant gaps between acquisition scores, the sketched and full-space rules select the same ordered batch. Thus, beyond a sufficient sketch dimension, increasing $p$ need not change the selected configurations, consistent with our experiments (Table~\ref{tab:sketch-ablation}). Full proofs are provided in Appendix~\ref{app:sketch-proof}.

\subsection{Sequential acquisition with rank-one updates}
\label{sec:methods-rank-one-updates}

Common uncertainty-based active-learning (AL) strategies, such as query-by-committee (QBC), score candidate configurations using uncertainty estimates from the current model. Selecting the top-$B$ configurations according to these fixed scores does not account for redundancy among selected points, since acquiring an informative configuration can reduce the uncertainty of nearby configurations. This issue is particularly relevant for trajectory-generated candidate pools, where configurations are often strongly correlated. Accounting for this reduction after each acquisition would require updating the uncertainty estimate sequentially, which for conventional neural-network uncertainty methods generally entails repeated model fitting or inference and can be computationally expensive.

We address this limitation with a greedy acquisition strategy that updates uncertainty after each selected configuration without retraining the network. Our approach builds on the Mahalanobis characterization of predictive uncertainty in NTK feature space (Section~\ref{sec:methods-theory}), which provides a geometric description of how uncertainty changes as configurations are incorporated. Directly recomputing this quantity after every selection would require repeated inversion of the feature covariance matrix, with \(\mathcal{O}(p^3)\) cost per update. Instead, we exploit the rank-one structure induced by each newly selected configuration and derive an efficient recursive update using the Sherman--Morrison identity.

Let $\mathbf{C}_k = \mathbf{F}_k + \lambda \mathbf{I}_P$ denote the regularized feature covariance at iteration $k$. The predictive uncertainty for a query $\mathbf{x}$ is estimated as $U_k(\mathbf{x}) = \lambda \phi(\mathbf{x})^\top \mathbf{C}_k^{-1} \phi(\mathbf{x})$, where $\lambda$ is calibrated on the validation set. Upon acquiring a new point $\mathbf{x}_{new}$, the covariance updates as
\[
    \mathbf{C}_{k+1} = \mathbf{C}_k + \phi(\mathbf{x}_{new}) \phi(\mathbf{x}_{new})^\top.
\]

Applying the Sherman--Morrison identity yields:
\begin{equation}\label{eq:sherman-morrison-update}
    \mathbf{C}_{k+1}^{-1} = \mathbf{C}_k^{-1} - \frac{\mathbf{C}_k^{-1} \phi(\mathbf{x}_{new}) \phi(\mathbf{x}_{new})^\top \mathbf{C}_k^{-1}}{1 + \phi(\mathbf{x}_{new})^\top \mathbf{C}_k^{-1} \phi(\mathbf{x}_{new})}.
\end{equation}

Substituting into the definition of $U_{k+1}(\mathbf{x})$ gives the recursive update:
\begin{equation}\label{eq:uncertainty-rank-one-update}
    U_{k+1}(\mathbf{x}) = U_k(\mathbf{x}) - \frac{\widetilde U_k(\mathbf{x}, \mathbf{x}_{new})^2}{\lambda + U_k(\mathbf{x}_{new})},
\end{equation}
where $\widetilde U_k(\mathbf{x}, \mathbf{x}_{new}) = \lambda \phi(\mathbf{x})^\top \mathbf{C}_k^{-1} \phi(\mathbf{x}_{new})$.

This yields an \(\mathcal{O}(p^2)\) update for sequential active learning, where uncertainty is reduced according to the squared geometric correlation between the query and the newly acquired configuration. Updating \(\mathbf{C}_k^{-1}\) costs \(\mathcal{O}(p^2)\). Once \(\mathbf{C}_k^{-1}\widetilde\phi(\mathbf{x}_{\mathrm{new}})\) is computed, rescoring a pool of \(m\) remaining configurations requires only \(\mathcal{O}(mp)\).

\section{Experiments}
\label{sec:experiments}
We first demonstrate that NTK-based uncertainty identifies large prediction errors, retains low-error predictions under selective retention, and can be calibrated. We then use AdaptNTK to show that this uncertainty remains effective when used as a sequential acquisition rule, followed by analyses of sketch size and computational cost. All methods use the same MACE architecture and training loss. For the active-learning comparisons, we use a fixed data split and random seeds shared across methods. Full experimental protocols are provided in Appendix~\ref{app:experimental-details}.

\subsection{Uncertainty quantification}
\label{sec:uncertainty-quantification}
To evaluate uncertainty quality, we compare the NTK-based uncertainty with standard approaches based on deep ensembles, Monte Carlo dropout (MCD), evidential deep learning (EDL), and stochastic weight averaging-Gaussian (SWAG) on held-out configurations from the rMD17 dataset. For each method, the estimated uncertainty is compared with the per-configuration force RMSE of the corresponding model. Table~\ref{tab:uq-metrics} reports error correlation, error-retention performance, and post-hoc calibration.

\begin{table}[htbp]
    \centering
    \caption{Performance of uncertainty-quantification methods on held-out configurations from the rMD17 dataset. Metrics are computed against each method's corresponding model and averaged across molecules. AURC$_n$ is normalized by oracle and random retention orderings, and ENCE is computed after out-of-fold recalibration; both are therefore comparable across rows. Bold indicates the best value in each column.}
    \label{tab:uq-metrics}
    \small
    \setlength{\tabcolsep}{4pt}
    \begin{tabular}{lcccc}
        \toprule
        Method & Spearman $(\rho_s)$ & Pearson $(\rho_p)$ & AURC$_n$ $\downarrow$ & ENCE $\downarrow$ \\
        \midrule
        NTK & $\mathbf{0.683 \pm 0.018}$ & $\mathbf{0.706 \pm 0.021}$ & $0.312 \pm 0.014$ & $0.0125 \pm 0.0018$ \\
        Ensemble              & $0.642 \pm 0.022$ & $0.661 \pm 0.024$ & $\mathbf{0.310 \pm 0.013}$ & $0.0153 \pm 0.0021$ \\
        MCD            & $0.348 \pm 0.031$ & $0.330 \pm 0.034$ & $0.633 \pm 0.027$ & $\mathbf{0.0105 \pm 0.0015}$ \\
        EDL           & $0.302 \pm 0.028$ & $0.262 \pm 0.032$ & $0.735 \pm 0.031$ & $0.0238 \pm 0.0034$ \\
        SWAG                  & $0.463 \pm 0.026$ & $0.471 \pm 0.029$ & $0.547 \pm 0.024$ & $0.0476 \pm 0.0052$ \\
        \bottomrule
    \end{tabular}
\end{table}

The NTK-based uncertainty shows the strongest linear and rank association with prediction error (\(\rho_p=0.706\), \(\rho_s=0.683\)). Its rank correlation is slightly higher than that of the three-member ensemble (\(0.642\)), while its AURC\(_n\) is comparable (\(0.312\) versus \(0.310\)). These results indicate that a single trained network can provide uncertainty discrimination comparable to an ensemble. The differences relative to SWAG, MCD, and EDL are larger for both correlation and error-retention performance.

\begin{figure}[!t]
    \centering
    \begin{subfigure}[t]{0.31\textwidth}
        \centering
        \includegraphics[width=\linewidth]{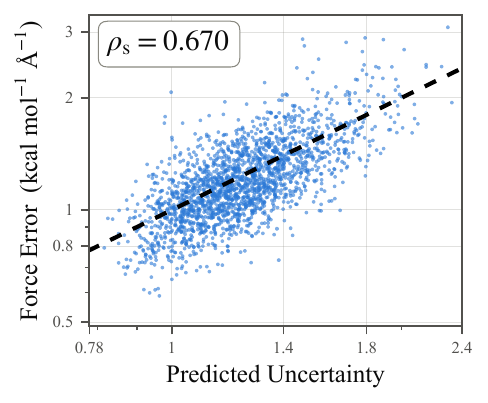}
        \caption{NTK}
        \label{fig:uq-ntk}
    \end{subfigure}
    \begin{subfigure}[t]{0.31\textwidth}
        \centering
        \includegraphics[width=\linewidth]{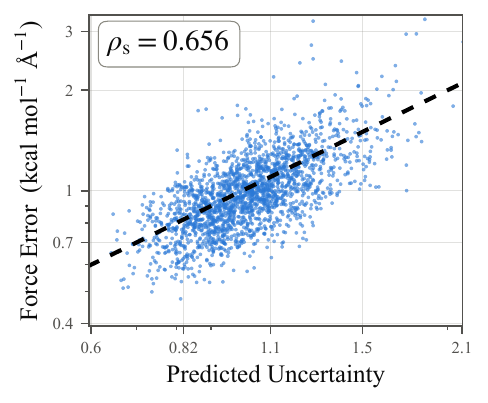}
        \caption{Ensemble}
        \label{fig:uq-ensemble}
    \end{subfigure}
    \begin{subfigure}[t]{0.31\textwidth}
        \centering
        \includegraphics[width=\linewidth]{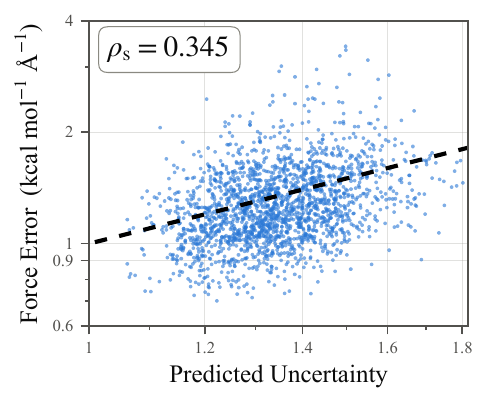}
        \caption{MC dropout}
        \label{fig:uq-mcd}
    \end{subfigure}
    \medskip
    \begin{subfigure}[t]{0.31\textwidth}
        \centering
        \includegraphics[width=\linewidth]{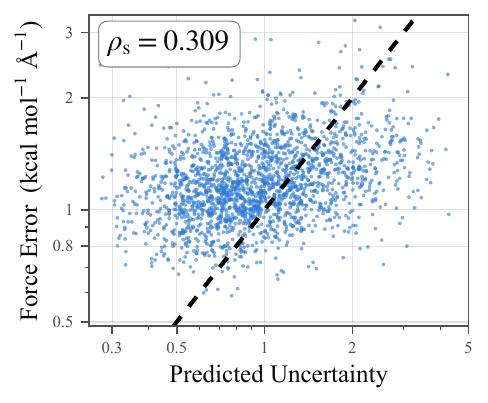}
        \caption{Evidential}
        \label{fig:uq-edl}
    \end{subfigure}
    \begin{subfigure}[t]{0.31\textwidth}
        \centering
        \includegraphics[width=\linewidth]{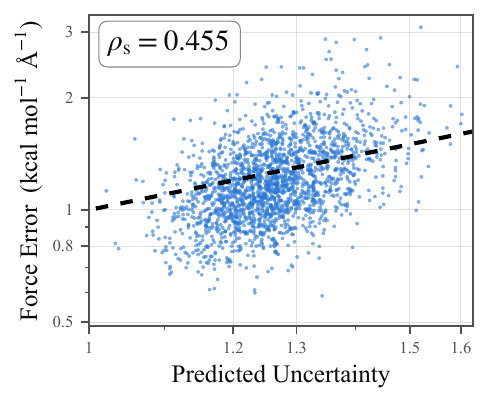}
        \caption{SWAG}
        \label{fig:uq-swag}
    \end{subfigure}
    \caption{Per-configuration force error versus predicted uncertainty on held-out aspirin configurations from rMD17. Each point is one
configuration; both axes are logarithmic and in kcal~mol$^{-1}$~\AA$^{-1}$, the
uncertainty having been recalibrated onto the error scale, so the dashed line
marks $y=x$. Insets give the Spearman correlation $\rho_s$, which is invariant to calibration. Panel~(d) uses a scale-only calibration (Appendix~\ref{sec:appendix-recalibration}), and each method is scored against its own trained model.}
\label{fig:uq-scatter}
\end{figure}

Calibration and discrimination measure different properties. MCD achieves the best ENCE after two-parameter recalibration, yet its \(\rho_s=0.348\) and AURC\(_n=0.633\) indicate substantially weaker ranking performance. A low ENCE can therefore coexist with limited acquisition utility, since recalibration can correct the overall uncertainty scale but does not alter the ranking. The NTK-based uncertainty is the only method that ranks among the top two for both discrimination metrics and ENCE. Figure~\ref{fig:uq-scatter} further illustrates that the NTK-based uncertainty and the ensemble maintain a broader and more ordered spread of uncertainty values.

\subsection{Comparison of data efficiency}
\label{sec:data-efficiency}

\begin{figure}[!htbp]
    \centering
    \includegraphics[width=\linewidth]{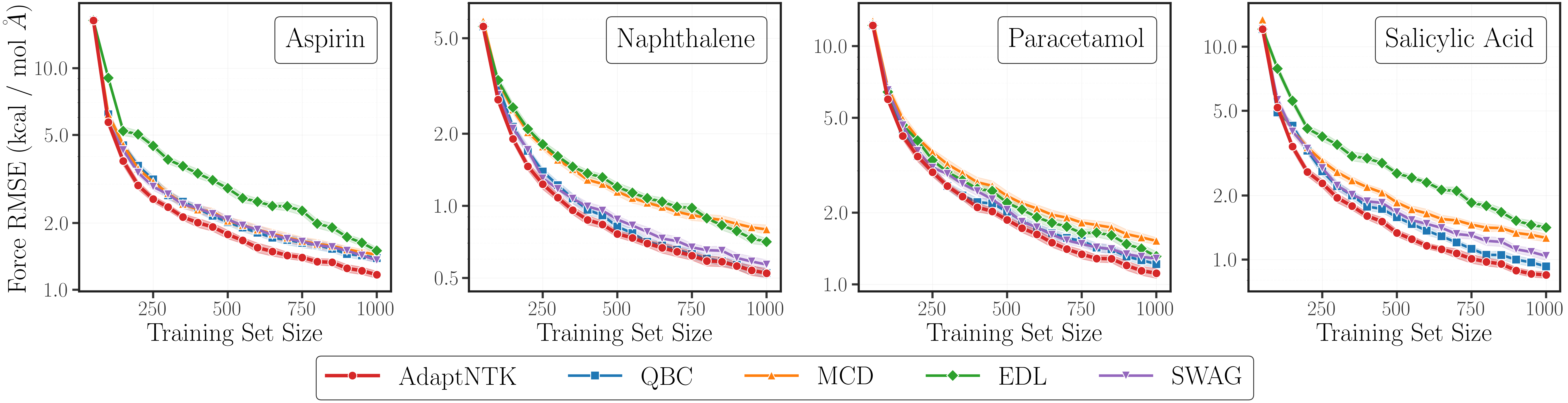}
    \caption{Active-learning performance for atomic force prediction. AdaptNTK is compared with established active-learning baselines (QBC, MCD, EDL, and SWAG) for four molecules from the rMD17 dataset. The y-axis (log scale) shows force RMSE (kcal\,mol$^{-1}$\,\AA$^{-1}$) on a holdout set as the training-set size increases. Shaded areas represent standard deviations in force errors averaged across three runs. AdaptNTK achieves faster convergence and lower absolute error than the baseline methods in all cases.}
    \label{fig:rmd17-learning-curves}
\end{figure}

\subsubsection{Active learning on rMD17}
\label{sec:rmd17-active-learning}
We next evaluate AdaptNTK's greedy sequential updates for active-learning acquisition. For each rMD17 molecule, 25,000 configurations are held out for testing, 50 fixed configurations initialize the training set, and the remaining configurations form the candidate pool. We report 20 checkpoints, including the initial model, as the labeled set grows from 50 to 1,000 configurations in increments of 50. All methods use the same fixed seeds, data split, architecture, and optimization schedule.
Across acquisition budgets, AdaptNTK attains the lowest mean force RMSE for aspirin, naphthalene, paracetamol, and salicylic acid (Figure~\ref{fig:rmd17-learning-curves}). Its advantage is largest at small and intermediate label budgets, where accounting for redundancy among selected configurations is most beneficial. The gap narrows at 1,000 labels, as all methods obtain broader coverage of the candidate pool.

\subsubsection{Reactive configurations in Transition-1X}
\label{sec:t1x-active-learning}
The rMD17 candidate pool and test set are drawn from the same molecular trajectories. To evaluate whether AdaptNTK can identify rare reactive configurations and support generalization across diverse molecules, we test performance on the Transition-1X dataset. We report results on both the general test split and a transition-state split, which isolates configurations near chemical transitions whose errors can be obscured when averaged over the broader test distribution.

\begin{figure}[htbp]
    \centering
    \begin{subfigure}[t]{0.48\textwidth}
        \centering
        \includegraphics[width=0.8\linewidth]{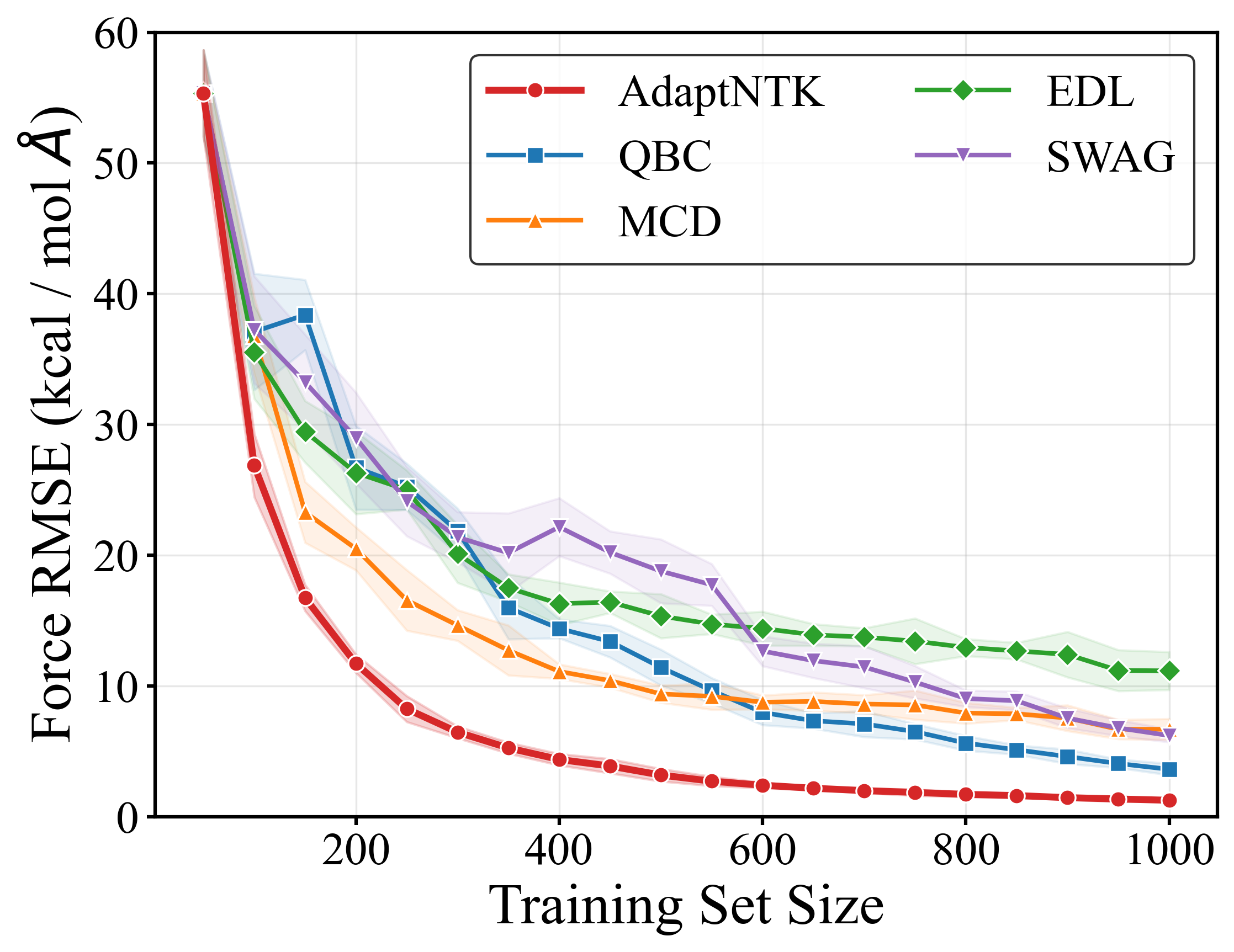}
        \caption{Learning curves}
        \label{fig:t1x-force-curves}
    \end{subfigure}\hfill
    \begin{subfigure}[t]{0.48\textwidth}
        \centering
        \includegraphics[width=0.8\linewidth]{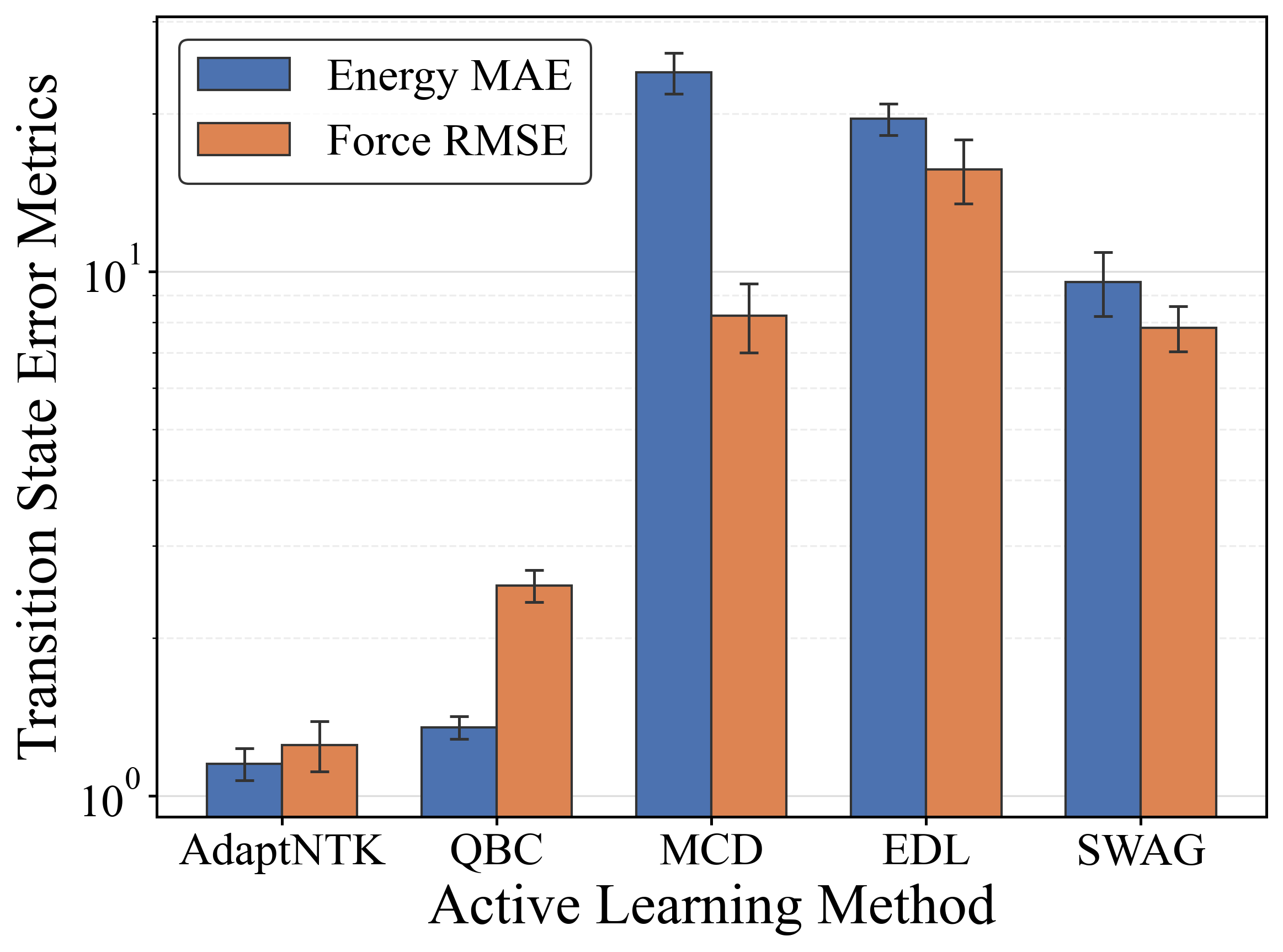}
        \caption{Transition-state metrics}
        \label{fig:t1x-transition-metrics}
    \end{subfigure}
    \caption{Active-learning performance on Transition-1X. AdaptNTK outperforms standard uncertainty-based active-learning methods, reaching comparable accuracy with less than half the labeled data required by the competing approaches. Transition-state error metrics are shown at 1,000 training configurations. The logarithmic vertical axes report force RMSE in \forceunit \ and energy MAE in \energyunit\ on the held-out test set.}
    \label{fig:t1x-results}
\end{figure}

AdaptNTK reduces force error most rapidly across the acquisition budget (Figure~\ref{fig:t1x-force-curves}). It reaches a force RMSE of approximately \(4\)~kcal mol\(^{-1}\)\AA\(^{-1}\) with about 450 labeled configurations; QBC requires close to 1,000, while the remaining baselines do not reach this level within the displayed budget. At 1,000 configurations, AdaptNTK reaches approximately \(1\)~kcal mol\(^{-1}\) \AA\(^{-1}\), compared with roughly \(4\) for QBC, \(6\)--\(7\) for MCD and SWAG, and \(11\) for EDL. Performance on the transition-state split shows the same ordering for both energy and force errors (Figure~\ref{fig:t1x-transition-metrics}). Values quoted here are rounded from the plotted curves for readability; all evaluations use the corresponding unrounded values.

\subsubsection{Ablations}
\label{sec:ablations}
To assess the contribution of recursive uncertainty updates, we compare AdaptNTK with a non-adaptive top-\(B\) baseline that scores the candidate pool once per 50-configuration batch and does not update uncertainty after selection. Sequential AdaptNTK performs consistently better throughout acquisition on Transition-1X, reaching a force RMSE of approximately \(1.2\) versus \(3\)~kcal mol\(^{-1}\) \AA\(^{-1}\) for the non-adaptive baseline (Figure~\ref{fig:topb-ablation} in Appendix~\ref{app:ablations}). The non-adaptive variant also consistently outperforms QBC at nearly all acquisition budgets, reflecting the stronger correlation between NTK uncertainty and prediction error. Random acquisition performs worse than both variants and QBC throughout the acquisition budget.

We also vary the sketch dimension from \(p=128\) to \(2048\). Final force RMSE ranges from \(1.22\) to \(1.34\)~kcal mol\(^{-1}\) \AA\(^{-1}\), with overlapping one-standard-deviation intervals and no monotonic improvement at larger sketch dimensions (Table~\ref{tab:sketch-ablation} in Appendix~\ref{app:ablations}), consistent with Corollary~\ref{cor:adaptntk-decision-stability}.
\subsection{Computational overhead}
\label{sec:computational-overhead}
AdaptNTK completes the reported timing run in 2,423~s, compared with 6,225~s for QBC, corresponding to a $2.57\times$ reduction in total runtime. This saving is driven primarily by model training and uncertainty evaluation: training a single model and computing its tangent features requires 1,748~s, whereas training the three-member QBC ensemble and evaluating all three models on every candidate structure requires 6,063~s. The label-free rank-one updates require no additional model fitting, and sequentially rescoring a candidate pool of approximately 75,000 structures after each selection requires only 676~s. Despite performing these adaptive updates throughout acquisition, AdaptNTK remains computationally competitive with other single-model approaches: its total runtime is within 6\% of MCD and 2\% of SWAG, while substantially reducing the cost relative to QBC. Combined with the stronger learning curves in Figures~\ref{fig:rmd17-learning-curves} and~\ref{fig:t1x-force-curves}, these results show that AdaptNTK achieves improved data efficiency while retaining the computational advantages of a single-model approach.

\begin{table}[htbp]
    \centering
    \caption{Computational overhead of active-learning methods. Execution time (in seconds) for pretraining and uncertainty quantification (UQ), batch selection, and the total active-learning cycle on the Transition-1X dataset.}
    \label{tab:execution-times}
    \small
    \begin{tabular}{lccc}
        \toprule
        Method & Pretraining + UQ (s) & Batch selection (s) & Total (s) \\
        \midrule
        AdaptNTK & 1747.50 & 675.84 & 2423.34 \\
        QBC      & 6062.56 & 162.59 & 6225.15 \\
        MCD      & 2144.71 & 144.60 & 2289.31 \\
        EDL      & 1013.73 & 113.72 & 1127.45 \\
        SWAG     & 2214.60 & 178.63 & 2393.23 \\
        \bottomrule
    \end{tabular}
\end{table}

\section{Limitations}
\label{sec:limitations}
The experiments have several important limitations. First, we evaluate uncertainty quality and active-learning performance for gas-phase molecules; extending this analysis to periodic systems is an important next step. Second, the uncertainty is constructed from energy-gradient features, which relate to force errors only indirectly. Features based on force-parameter Jacobians would provide a more direct representation, but are substantially larger. Third, force errors can be poor indicators of downstream simulation performance, and the relationship between uncertainty in predicted forces and uncertainty in simulation observables remains to be characterized. Finally, our active-learning comparisons focus on uncertainty-based methods and do not include dedicated diversity-based or chemistry-informed acquisition strategies. Diversity could, for example, be incorporated after uncertainty filtering by clustering or selecting diverse structures among the most uncertain configurations. Thus, our experiments demonstrate that NTK-based uncertainty provides an effective signal for active learning, rather than establishing AdaptNTK as superior to acquisition strategies that incorporate chemical intuition or other post-hoc selection heuristics.

\section{Conclusion}
\label{sec:conclusion}
The NTK-based uncertainty converts a trained neural potential into a post-hoc, pointwise uncertainty estimate without a committee or learned uncertainty head. It achieves the strongest error correlations across the four rMD17 molecules and nearly matches the three-member ensemble in error retention. AdaptNTK provides the downstream active-learning demonstration: the same uncertainty can be updated within an acquisition batch without reference labels or additional model fitting, and achieves the lowest reported errors in the rMD17 and Transition-1X studies. The top-\(B\) ablation supports the value of sequential updates, while the sketch-dimension sweep shows no observed benefit beyond \(p=128\). Comparisons with broader acquisition baselines and extensions to periodic systems remain important directions for future work.

\section*{Acknowledgments}
We acknowledge startup funding from Cornell University and computing resources provided by the Cornell University Center for Advanced Computing (CAC). 

\bibliographystyle{plainnat}
\bibliography{references}

\clearpage
\appendix

\section*{Supplementary Information}
\addcontentsline{toc}{section}{Supplementary Information}
Appendix~\ref{app:sketch-proof} gives the complete proof of the Gaussian-sketch theorem and its active-learning corollary. Appendix~\ref{app:ablations} separates the effect of sequential score updates from that of the sketch dimension. Appendix~\ref{app:experimental-details} records the data splits, active-learning protocol, model and training settings, uncertainty baselines, evaluation metrics, calibration procedure, and timing protocol.
\section{Proof of the sketch-preservation result}
\label{app:sketch-proof}

This appendix proves Theorem~\ref{thm:ntk-sketch-preservation} and
Corollary~\ref{cor:adaptntk-decision-stability}.

Fix $x\in\mathcal X$, abbreviate $Z=Z_x$, and set
\[
K=ZZ^\top,\qquad M=K+\lambda I,\qquad H=M^{-1/2}Z.
\]
Writing the rows of $S$ as $g_j^\top/\sqrt p$ with
$g_j\sim\mathcal N(0,I_P)$ gives
\begin{align}
M^{-1/2}(ZS^\top SZ^\top-K)M^{-1/2}
&=H(S^\top S-I_P)H^\top \nonumber\\
&=\frac1p\sum_{j=1}^p(Hg_j)(Hg_j)^\top-HH^\top .
\label{eq:ridge-covariance}
\end{align}
The vectors $Hg_j$ are Gaussian with covariance
$\Sigma=HH^\top=M^{-1/2}KM^{-1/2}$.  Its eigenvalues lie in $[0,1]$, and
\[
\operatorname{tr}(\Sigma)
=\operatorname{tr}\!\left[K(K+\lambda I)^{-1}\right]
=:d_\lambda(Z).
\]
Gaussian sample-covariance concentration in effective
dimension~\citep{koltchinskii2017covariance} therefore implies, for an absolute constant
$C$, that
\[
p\ge C\epsilon^{-2}\!\left(d_\lambda(Z)+\log\frac1\eta\right)
\quad\Longrightarrow\quad
\left\|H(S^\top S-I_P)H^\top\right\|_2\le\epsilon
\]
with probability at least $1-\eta$.  Taking
$\eta=\delta/m$ and a union bound yields this event simultaneously for all
$x\in\mathcal X$. Equivalently, with
$\widetilde K=ZS^\top SZ^\top$ and $\widetilde M=\widetilde K+\lambda I$,
\begin{equation}
(1-\epsilon)M\preceq\widetilde M\preceq(1+\epsilon)M.
\label{eq:regularized-spectral}
\end{equation}

It remains to translate \eqref{eq:regularized-spectral} into a score bound. The
rows of $Z$ except the last are $\Phi$; write
\[
K=\begin{bmatrix}K_0&k_x\\k_x^\top&\kappa_x\end{bmatrix},
\quad K_0=\Phi\Phi^\top,\quad k_x=\Phi\phi(x),\quad
\kappa_x=\|\phi(x)\|_2^2.
\]
The Woodbury identity gives
\[
\kappa_x-k_x^\top(K_0+\lambda I)^{-1}k_x
=\lambda\phi(x)^\top(\Phi^\top\Phi+\lambda I_P)^{-1}\phi(x)
=U(x).
\]
Hence the Schur complement of the upper-left block of $M$ is
$s=\lambda+U(x)$; the corresponding sketched Schur complement is
$\widetilde s=\lambda+\widetilde U(x)$.
Inverting \eqref{eq:regularized-spectral}, taking its last diagonal entry, and
using $(M^{-1})_{-1,-1}=s^{-1}$ yields
\[
(1-\epsilon)s\le\widetilde s\le(1+\epsilon)s.
\]
After subtracting $\lambda$, this is exactly
\[
|\widetilde U(x)-U(x)|\le\epsilon(\lambda+U(x)),
\]
which proves Theorem~\ref{thm:ntk-sketch-preservation}. \hfill $\square$

For Corollary~\ref{cor:adaptntk-decision-stability}, fix the ordered greedy path obtained
with the full features. At step $t$, augment $\Phi$ with the previously selected
features and apply the preceding argument to every remaining candidate. A union
bound over at most $Bm$ step--candidate pairs, together with the definition of
$d_\lambda^{\rm AL}$, gives the same score bound everywhere along this fixed
path. Let $x_t^\star$ be its maximiser. Then
\begin{align*}
\widetilde U_{A_t}(x_t^\star)&\ge
U_t^{(1)}-\epsilon(\lambda+U_t^{(1)}),\\
\max_{x\ne x_t^\star}\widetilde U_{A_t}(x)&\le
U_t^{(2)}+\epsilon(\lambda+U_t^{(2)}).
\end{align*}
The first quantity is strictly larger whenever the stated normalised gap
exceeds $\epsilon$.  Induction on $t$ then shows that the sketched procedure
follows the full-space ordered path. \hfill $\square$

\section{Ablation studies}
We report two controlled analyses: one isolates the contribution of the sequential update, and the other tests sensitivity to the Gaussian-sketch dimension.
\label{app:ablations}
\begin{figure}[htbp]
    \centering
    \includegraphics[width=0.45\linewidth]{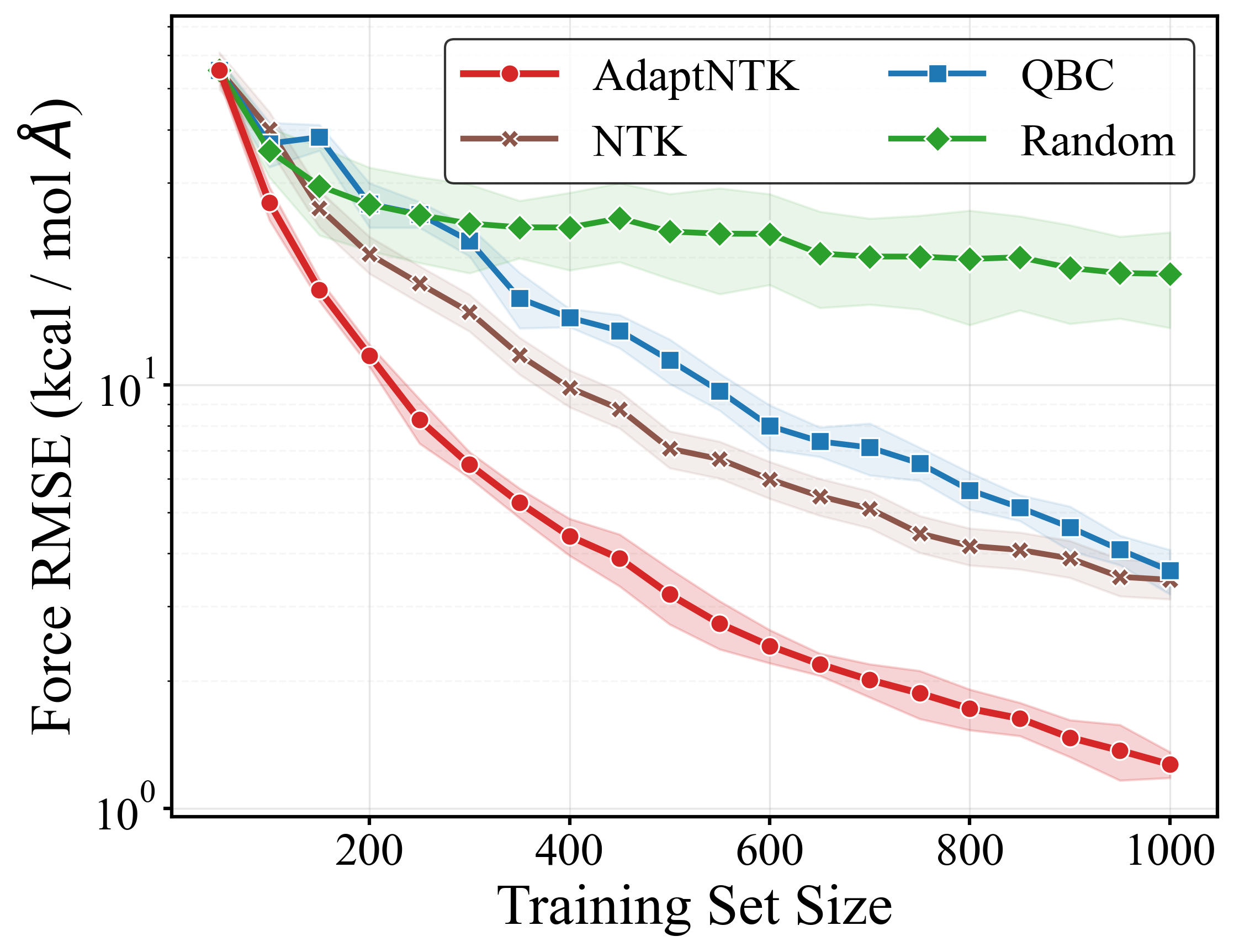}
    \caption{Sequential-update ablation on Transition-1X.
    AdaptNTK updates all remaining scores after each selection. ``NTK'' denotes
    the non-adaptive top-$B$ ablation: it uses the same NTK score but selects the
    50 highest-scoring configurations without within-batch updates. QBC and Random baselines are
    included as context. All curves use the same initial 50 configurations,
    fixed transition-state-only test split, fixed seeds and 20-checkpoint budget
    ending at 1,000 labels. Shaded areas represent standard deviation in force errors averaged across 3 runs.}
    \label{fig:topb-ablation}
\end{figure}

\begin{table}[htbp]
    \centering
    \caption{Gaussian-sketch ablation on Transition-1X. Final force
    RMSE on the test split after the same 20-checkpoint
    active-learning protocol, with every setting ending at $N=1{,}000$.}
    \label{tab:sketch-ablation}
    \small
    \begin{tabular}{rcc}
        \toprule
        Sketch dimension $p$ & Force RMSE
        (kcal\,mol$^{-1}$\,\AA$^{-1}$) \\
        \midrule
        128  & $1.22 \pm 0.10$ \\
        256   & $1.23 \pm 0.08$ \\
        512  & $1.27 \pm 0.12$ \\
        1,024 & $1.34 \pm 0.10$ \\
        2,048  & $1.32 \pm 0.08$ \\
        \bottomrule
    \end{tabular}
\end{table}

The maximum difference among point estimates is
$0.12$~kcal\,mol$^{-1}$\,\AA$^{-1}$, and the ordering is non-monotone. Thus the
reported run provides no evidence that a larger sketch improves final force
error beyond $p=128$. This empirical plateau is consistent with
Corollary~\ref{cor:adaptntk-decision-stability}: acquisition becomes insensitive to $p$
once sketch error is below the relevant score gaps.

\section{Experimental details}
\label{app:experimental-details}

\subsection{Datasets and splits}
\label{sec:appendix-datasets}
Uncertainty-quality experiments use four rMD17 molecular datasets: aspirin, naphthalene, paracetamol, and
salicylic acid. Each dataset contains configurations sampled from an \textit{ab initio} molecular-dynamics trajectory. Energies are reported in kcal mol\(^{-1}\) and forces in kcal mol\(^{-1}\) \AA\(^{-1}\). For each molecule, we draw 1{,}000 training configurations and 100 validation configurations. From the remaining configurations, we draw disjoint fixed evaluation and calibration subsets of 2{,}000 configurations each. All partitions use the same fixed seeds, every method is evaluated on identical configurations, and we verify disjointness by hashing coordinates.

For active learning, 25{,}000 configurations are held out once for
testing, 50 fixed configurations initialize training, and every remaining
configuration forms the initial acquisition pool. The 20 plotted checkpoints
include this initial fit and then add 50 configurations at a time, ending at 1{,}000 labeled configurations.

For Transition-1X, an additional evaluation set is constructed from the
dataset's transition-state split. These configurations are excluded from
pretraining. This deliberately tests whether the acquisition
strategy improves the region of chemical space most relevant to reactivity,
rather than allowing the metric to be dominated by more numerous non-transition
configurations. The learning curves use the same fixed seeds and are reported through 1{,}000 labeled
configurations.

\subsection{Active-learning loop}
\label{sec:appendix-active-learning-loop}
At every checkpoint, the current training set is used to fit the potential and
the method-specific uncertainty estimator. Each method then selects the next 50
configurations from its remaining pool. AdaptNTK constructs the batch
sequentially: after each choice it applies
Eq.~\eqref{eq:uncertainty-rank-one-update} to the remaining
scores, but does not use the selected configuration's label. The baselines rank
the pool under their fitted uncertainty and take the highest-scoring batch. Once
the batch is complete, its dataset labels are revealed, appended to the training
set, and the potential is retrained. The data split, initial configurations,
random seed, architecture, label budget, and evaluation set are held fixed
across methods; only the acquisition rule changes.

For the top-$B$ ablation, the same NTK-based score is evaluated once at the beginning
of each acquisition batch and the 50 largest values are selected without
applying Eq.~\eqref{eq:uncertainty-rank-one-update}. For the sketch ablation,
only $p$ changes; the data,
model, regularization, acquisition budget, and fixed seed are unchanged.

\subsection{Interatomic potential and training}
\label{sec:appendix-model-training}
All methods use the same MACE architecture~\citep{batatia2022mace}
(\texttt{ScaleShiftMACE}): a 6.0\,\AA{} cutoff, hidden irreps
$128\times0e + 128\times1o$, two interaction layers, correlation order 3, 8
Bessel radial basis functions, polynomial cutoff order 5, and $\ell_{\max}=3$,
giving 724{,}240 parameters. Models are trained for 500 epochs with AdamW at a
constant learning rate of $10^{-4}$ and batch size 16, minimizing
$0.01\,\mathcal{L}_{E} + 0.99\,\mathcal{L}_{F}$ where both terms are mean squared
errors. The checkpoint with the lowest validation loss is retained. Energy shift
and scale are set from the per-atom training energy statistics.

\subsection{Uncertainty methods}
\label{sec:appendix-uncertainty-methods}
Each method reduces a configuration to a single scalar uncertainty $u$.

\paragraph{NTK-based uncertainty:} Energy gradients $\phi(x)=\nabla_\theta E(x;\theta)$ are
extracted from the trained model and projected to $p=512$ dimensions with a fixed
Gaussian sketch (Theorem~\ref{thm:ntk-sketch-preservation}). The score is the regularized Mahalanobis form
$U(x)=\lambda\,\tilde\phi(x)^\top(\tilde F+\lambda I_p)^{-1}\tilde\phi(x)$, where $\tilde F$ is the sketched feature covariance of the
training set.

\paragraph{Ensemble (QBC):} Three models are trained with identical settings, but
different random seeds. The score is the standard deviation of predicted force
components across members, averaged over atoms and Cartesian directions. The
error target for this method uses the ensemble-mean forces.

\paragraph{MCD:} The model is trained with dropout ($p=0.1$), applied after the non-linearity of each readout block, and dropout remains active at inference time. The uncertainty score is the standard deviation of the predicted forces across 20 stochastic forward passes.

\paragraph{SWAG:} Weight snapshots are collected over the final 25\% of training
epochs (at most 20 retained) to form a Gaussian posterior with diagonal plus
low-rank covariance. The score is the across-sample standard deviation of
predicted forces over 30 posterior draws.

\paragraph{Evidential:} A normal-inverse-gamma head is fitted post hoc on the
frozen backbone, mapping mean-pooled invariant ($\ell=0$) node features to
$(\gamma,\nu,\alpha,\beta)$ by minimizing the NIG negative log-likelihood with
evidence regularization. Targets and features are standardised before fitting;
without this the likelihood is minimized by $\nu\to0$, which inflates the
epistemic term by many orders of magnitude while leaving it nearly constant
across configurations. The score is the epistemic variance
$\beta/(\nu(\alpha-1))$.

\subsection{Error target and metrics}
\label{sec:appendix-error-metrics}
For each configuration $i$ the error is the force RMSE over all atoms and
Cartesian components,
$e_i = \big[\tfrac{1}{3N}\sum_{a,c}(F^{\text{pred}}_{ac}-F^{\text{true}}_{ac})^2\big]^{1/2}$,
in kcal\,mol$^{-1}$\,\AA$^{-1}$. Each method is evaluated against the predictions
of the model it is defined on; consequently the mean error differs by method. Spearman and Pearson coefficients are computed between
$u$ and $e$.

\paragraph{AURC$_n$.} Configurations are ranked by $u$ and progressively removed
in decreasing order of uncertainty. The RMSE of the retained set is tracked
against the retained fraction, and AURC is the area under this curve. Because a
more accurate model lowers this area irrespective of uncertainty quality, we
report the normalised quantity
$\text{AURC}_n = (\text{AURC}-\text{AURC}_{\text{oracle}})/(\text{AURC}_{\text{random}}-\text{AURC}_{\text{oracle}})$,
where the oracle ranks by true error and the random baseline averages five
permutations, both computed from the same $e$. Thus $0$ denotes a perfect ranking
and $1$ no better than chance, and the measure is comparable across methods
whose models differ in accuracy.

\paragraph{Recalibration and ENCE.}
\label{sec:appendix-recalibration}
Raw scores are not commensurate with the
error: some are variances, some standard deviations, and the NTK-based score
carries an arbitrary scale through $\lambda$. We therefore fit
$\sigma^2 = a\,u + b^2$ by minimizing the $\chi^2$ negative log-likelihood
$\sum_i \log\sigma_i^2 + e_i^2/\sigma_i^2$, in which $a$ absorbs the unit
conversion and $b^2$ captures the irreducible error a purely epistemic term
cannot represent. Parameters are fitted by two-fold cross-fitting within the
scored set and applied only to the held-out fold, so the reported ENCE is not
self-graded. ENCE is the mean over ten equal-count $\sigma$ bins of
$|\text{RMV}-\text{RMSE}|/\text{RMV}$. For evidential regression, the fitted
floor $b=1.19$ is close to the mean error $1.23$ and compresses the display;
Figure~\ref{fig:uq-edl} therefore uses $\sigma^2=a\,u$ for display only.
This monotone change leaves ranks unchanged, while Table~\ref{tab:uq-metrics}
uses the two-parameter ENCE.

All timing measurements in Table~\ref{tab:execution-times} use one NVIDIA A100 GPU per method. The reported wall-clock components cover pretraining plus uncertainty construction and batch selection. They exclude the cost of obtaining reference labels, which is identical for a fixed acquisition count but is the quantity reduced when one method reaches a target error with fewer acquisitions.
\end{document}